\documentclass[11pt,a4paper]{article}
\usepackage[hyperref]{acl2021}
\usepackage{times}
\usepackage{latexsym}

\usepackage{caption}
\usepackage{subfig}
\usepackage{graphicx}
\usepackage{listings}
\usepackage{paralist}
\usepackage{amssymb}
\usepackage{amsmath}
\usepackage{tikz}
\usetikzlibrary{calc}
\newcommand{\tikzmark}[1]{\tikz[overlay,remember picture] \node (#1) {};}

\newcommand{\bfa}{\mathbf{a}}
\newcommand{\bfr}{\mathbf{r}}

\newcommand{\bfh}{\mathbf{h}}

\newcommand{\bfx}{\mathbf{x}}

\newcommand{\bfs}{\mathbf{s}}
\newcommand{\bfp}{\mathbf{p}}

\newcommand{\bR}{\mathbf{R}}
\newcommand{\bA}{\mathbf{A}}

\makeatletter
\newcommand*\bigcdot{\mathpalette\bigcdot@{.6}}
\newcommand*\bigcdot@[2]{\mathbin{\vcenter{\hbox{\scalebox{#2}{$\m@th#1\bullet$}}}}}
\makeatother

\lstdefinestyle{customc}{
  belowcaptionskip=1\baselineskip,
  breaklines=true,
  frame=L,
  xleftmargin=\parindent,
  language=C,
  showstringspaces=false,
  basicstyle=\small \ttfamily,
  keywordstyle=\bfseries\color{green!40!black},
  commentstyle=\itshape\color{purple!40!black},
  identifierstyle=\color{blue},
  stringstyle=\color{orange},
}
\DeclareCaptionFont{white}{ \color{white} }
\DeclareCaptionFormat{listing}{
  \colorbox[cmyk]{0.43, 0.35, 0.35,0.01 }{
    \parbox{\textwidth}{\hspace{15pt}#1#2#3}
  }
}

\usepackage{microtype}

\aclfinalcopy 

\title{Learning Slice-Aware Representations with Mixture of Attentions}

\author{Cheng Wang  \quad Sungjin Lee \quad
  Sunghyun Park \quad Han Li \quad Young-Bum Kim \quad Ruhi Sarikaya\\
  Amazon Alexa AI \\
  \texttt{\{cwngam,sungjinl,sunghyu,lahl,youngbum,rsarikay\}@amazon.com} \\}

\date{}

\begin{document}
\maketitle
\begin{abstract}
Real-world machine learning systems are achieving remarkable performance in terms of coarse-grained metrics like overall accuracy and F-1 score. However, model improvement and development often require fine-grained modeling on individual data subsets or slices, for instance, the data slices where the models have unsatisfactory results. In practice, it gives tangible values for developing such models that can pay extra attention to critical or interested slices while retaining the original overall performance. This work extends the recent slice-based learning (SBL)~\cite{chen2019slice} with a mixture of attentions (MoA) to learn slice-aware dual attentive representations. 
We empirically show that the MoA approach outperforms the baseline method as well as the original SBL approach on monitored slices with two natural language understanding (NLU) tasks.
\end{abstract}

\begin{table*}
\centering
\begin{tabular}{ll}
\begin{lstlisting}
def SF_Length(utterance,k=10):
    return len(utterance) < k

def SF_Time(utterance):
    return "time" in utterance

def SF_Email(utterance):
    return "email" in utterance
\end{lstlisting}
& 
\begin{lstlisting}
def SF_Long(sentence,k=10):
    n = len(sentence
            .split(' '))
    return  n > k

def SF_Question(sentence):
    return sentence[-1] 
            == '?'
\end{lstlisting}
\end{tabular}
\caption{The designed slice functions (SFs)\footnotemark. Left: We monitor three data slices - short utterances, those involving ``time", and those involving ``email". Right: We monitor long sentences and questions\footnotemark.}
\label{tab:slice}
\vspace{-5mm}
\end{table*}

\section{Introduction}
Though machine learning systems have been achieving excellent performance in terms of coarse-grained metrics like accuracy, they perform poorly or even fail on some individual data subsets (i.e., slices). For instance, many models have difficulties when learning for classes with only a few samples or samples with challenging structures.
\begin{figure}[!t]
\vspace{-5mm}
\centering
    \includegraphics[width=0.5\textwidth]{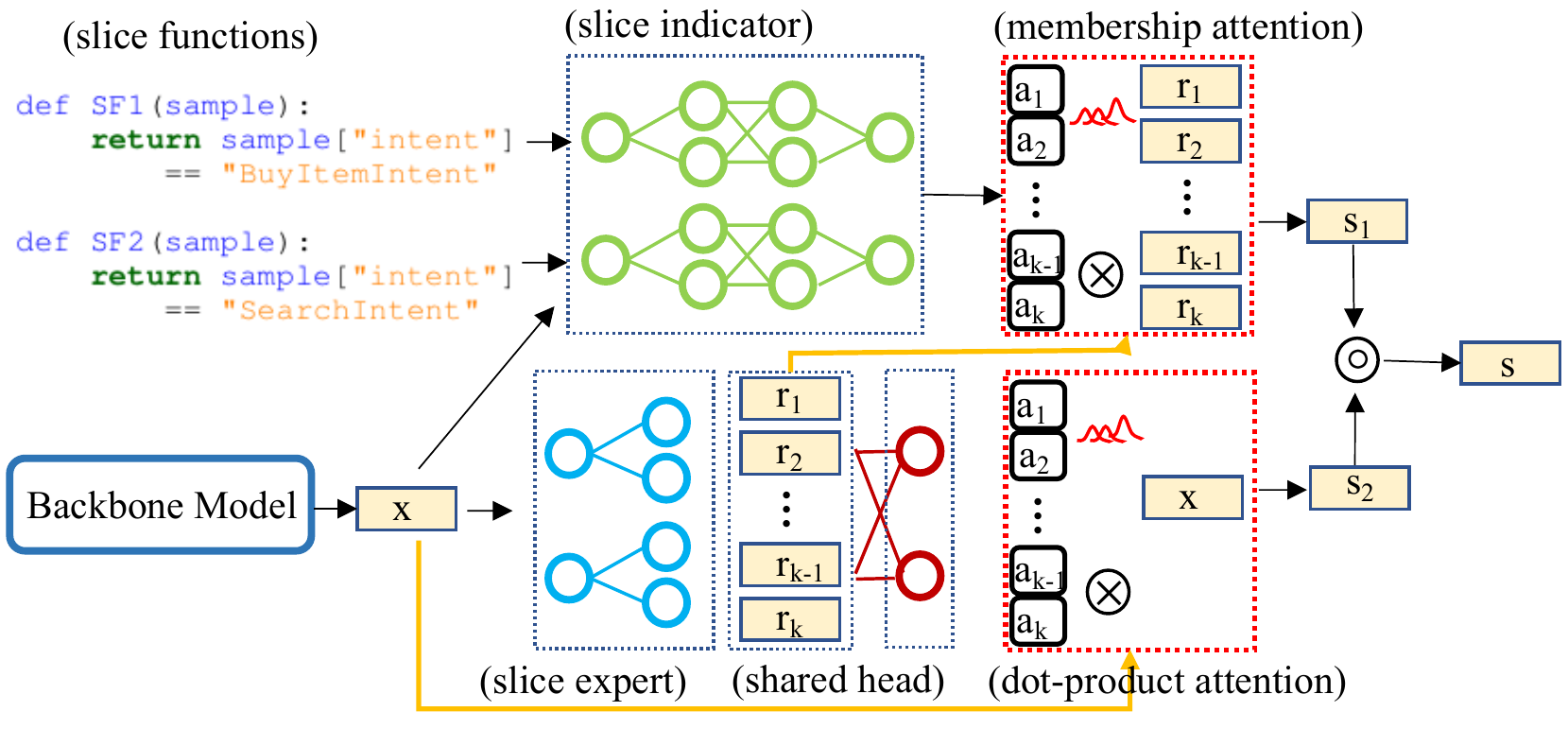} 
\caption{\label{fig:slice_model} The slice-aware architecture with MoA. It consists of six components: (1) \textit{slice functions} define the special data slices that we want to monitor; (2) \textit{backbone model} for feature extraction (e.g., BERT); (3) \textit{slice indicators} are membership functions to predict if a sample belongs to the slice; (4) \textit{slice experts} aim to learn slice-specific representations; (5) \textit{shared head} is the base task predictive layer across experts and (6) the proposed \textit{mixture of attentions (MoA)} learns to attend to the slices of interest. It contains two different attention mechanism (red boxes): a membership attention and a dot-product attention. The MoA learns to re-weight the expert representation $\bfr$ to a slice-aware representation $\bfs$ and the original representation $\bfx$ to $\bfs$ (yellow lines). The slice distributions are computed in deterministic (weighted sum of slices) or stochastic (sampling) way in re-weighting $\bfr$ and $\bfx$.}
\vspace{-2mm}
\end{figure}  
Inspecting particular data slices can serve as an important component in model development cycles. A recently proposed slice-based learning (SBL) exhibited compelling results with more than 3\% improvements on pre-defined slices~\cite{chen2019slice} in the task of binary classification. However, one potential limitation of the existing attention mechanism in SBL is that in multi-class cases, the attention suffers from the difficulty in using the experts' confidences appropriately for computing slice distributions (refer to Sec. \ref{sec:sadrm}). 

In this paper, we extend SBL with a mixture of attentions (MoA) mechanism.
Two different attention mechanisms are learned to jointly attend to the defined slices from different representations in different latent subspaces. The first attention is based on slice membership likelihood and/or experts confidence as in SBL~\cite{chen2019slice}, which we call \textit{membership attention}. The second one is \textit{dot-product attention} that is based on the backbone model (e.g., BERT~\cite{devlin2019bert}) extracted representations. The MoA approach is akin to multi-head attention~\cite{vaswani2017attention} but with different attention types that receive different inputs.

As presented in Figure \ref{tab:slice}, the two attentions in MoA can work jointly to attend to (1) the expert representation $\bfr$ and (2) the backbone model extracted representation $\bfx$, and finally form an attentive representation $\bfs$. The $\bfs$ is a slice-aware featurization of the samples in the particular data slices and will be used for making a final model prediction.

We argue that learning joint attention with MoA from different resources for computing slice distributions is beneficial~\cite{vaswani2017attention,li2018multi}. 
We evaluate the effectiveness of our proposed approach on intent detection~\cite{liu2019benchmarking} and linguistic acceptability~\cite{warstadt2018neural} tasks.

Our main contributions are twofold: 
\begin{itemize}
    \item We extend SBL with MoA. The MoA approach has the ability to attend to slices in deterministic (weighted summation) and stochastic (sampling) ways.
    \item We conduct extensive experiments on two NLU tasks. The results show that MoA outperforms the baseline and vanilla SBL by average up to 9\% and 6\% respectively on defined slices.
\end{itemize}

\addtocounter{footnote}{-2} 
\stepcounter{footnote}\footnotetext{The SFs are task-dependent and not assumed to be perfectly accurate. They can be noisy or from weak supervision sources~\cite{ratner2016data}. Here, for the task in sec.\ref{sec:linguistic_acceptability}, SFs are defined to improve the slices where the model has unsatisfactory results as compared to the overall performance. For the task in sec.\ref{sec:intent_classification}, we define the SFs for the slices of interest.}
\stepcounter{footnote}\footnotetext{Alternatively, 5W1H rule for questions~\cite{kim2019probing}.}

\section{Architecture}
\label{sec:slice-aware}
Figure \ref{fig:slice_model} presents the slice-aware architecture based on SBL~\cite{chen2019slice}. Let $\{x^n, y^n\}_n^N$ be a dataset with N samples. We aim to learn slice-aware representation $\bfs$ from slice-experts-learned representation $\bfr$ and backbone-model-extracted representation $\bfx$. 

We first define \textbf{slice functions (SFs)} as in Table~\ref{tab:slice} to split the dataset into $k$ slices 
of interests. Each sample is assigned with a slice label  $\gamma \in [0, 1]$ in $\{\gamma_1, \gamma_2,...,\gamma_k\}$ as supervision data\footnote{$s_1$ is the base slice, and $s_2$ to $s_k$ are the slices of interest.}.

Second, we use a \textbf{backbone model} like BERT to extract representation $\bfx \in \bR^d$ for a given sample.
Then, \textbf{slice indicators} $f_i(\bfx;\mathbf{w}^f_i)$, $\mathbf{w}^f_i\in \mathcal{R}^{d\times 1}$, $i\in\{1,..,k\}$ map $\bfx$ to a prediction $h_i$. $f_i$ are trained with $\{\bfx^{n},\gamma^{n}\}_{n}^N$ to predict whether a sample belongs to a particular slice. They are learned with cross entropy loss
\begin{equation}
\zeta_{1}=\sum_{i}^k \mathcal{L}_{CE}(h_i,\gamma_i)
\end{equation}

Then, \textbf{slice experts} $g_i(\bfx;\mathbf{w}^g_i)$, $\mathbf{w}^g_i \in \mathcal{R}^{d\times d}$ learn a mapping from $\bfx$ to a slice vector $ r_i \in \mathcal{R}^{d}$ with the samples that only belong to the slice, followed by a \textbf{shared head}, which is shared across all experts and maps $r_i$ to a prediction $\hat{y}=\varphi(r_i; \mathbf{w}_s)$. $g_i$ and $\varphi$ are learned on the base (original) task with ground-truth label $y$ by 
\begin{equation}
\zeta_2=\sum_{i}^k\gamma_i  \mathcal{L}_{CE}(\hat{y}, y) 
\end{equation}

Finally, a \textbf{mixture of attentions}(MoA) (as in Sec. \ref{sec:sadrm}) re-weights $\bfr$ and $\bfx$ to form $\bfs$. The $\bfs$ goes through a final prediction function $\eta$ on the base task. The loss function is 
\begin{equation}
\zeta_3=\mathcal{L}_{CE}(\eta(\bfs; \mathbf{w}_p), y)
\end{equation}

The total loss is a combination of the loss for slice indicators, slice experts and base task prediction function:
\begin{equation}
\zeta = \zeta_1 +\zeta_2 + \zeta_3
\end{equation}
The whole model is optimised with back-propagation~\cite{rumelhart1986learning} in an end-to-end way.

\section{Methodology}
\label{sec:sadrm}
The SBL approach~\cite{chen2019slice} proposed a slice-residual attention modules (SRAMs) that are directly based on stacked membership likelihood $H \in \bR^{k}$ and experts' prediction confidence $\left|Y\right|\in\bR^{c\times k}$, $c=1$ (i.e., binary classification). Then, slice distribution (attention weights) is computed with $a=\textsc{softmax}(H+\left | Y \right |)$. 
One potential limitation of this mechanism is that the above formulation can lead to mismatch shape in element-wise addition when $c>2$ (i.e., multi-class classification). To circumvent this, we propose a mixture of attentions (MoA) to augment membership attention with dot-product attention from different information resources.

\subsection{Mixture of Attentions}

Let $\bfx \in \bR^d$ be the original representation from the backbone model (e.g., BERT), $h_i \in \bR^c$ ($c=1$) as i-th indicator function's prediction, and $r_i \in \bR^d$ as i-th expert learned representation. When stacking on $k$ slices, we have $\bfh \in \bR^{c\times k}$ and $\bfr \in \bR^{d\times k}$. MoA's goal is to (1) attend to $\bfr$ based on indicator functions' membership likelihood and/or experts confidence\footnote{In multi-class case, only membership likelihood is used.}; (2) attend to $\bfx$ with a dot-product attention; (3) to form a new slice-aware attentive representation $\bfs\in \bR^{d}$ with weighted (sampled) $\bfr$ and $\bfx$.

The slice distributions
are computed differently. For membership attention, the probability $\bfp_1 = \textsc{softmax}(\mathbf{h})$ or $\bfp_1 = \textsc{softmax}(\mathbf{h}+\left | \bfr \right |) \in \bR^k$ ($d$=1 in binary classification). Then membership weighted slice representation is computed: $\bfs_1 = \bfr \bigcdot \bfp_1, \bfs_1\in \bR^d $. For dot-product attention, we aim to learn an attention matrix $\bA=\{\bfa_1,..., \bfa_k\}, \bfa \in \bR^d, \bA \in \bR^{d\times k}$ is randomly initialized and learned by the standard back-propagation. Intuitively, each $\bfa$ is learned to be a slice prototype~\cite{wang2019state,roy2020efficient}. The probability over slices is computed as:
\begin{equation}
\bfp_2 = \textsc{softmax}(\bA^\top\bigcdot \bfx) \in \bR^k
\end{equation}
A new attentive representation $\bfs_2$ is formed by weighting $\bA$ with $\bfp_2$:
\begin{equation}
\bfs_2 =  \bA  \bigcdot  \bfp_2, ~~~\bfs_2 \in \bR^d
\end{equation}
or sampling from $\bA$:
\begin{equation}
\mbox{sample } \bfs_2 \sim \{\bfa_1,..., \bfa_k\}
\end{equation}
Then slice-aware vector $\bfs$ is computed by
\begin{equation}
\bfs = \bfs_1 \circledcirc \bfs_2
\label{eq:att_residual}
\end{equation}
where $\circledcirc$ is an operator (either $\oplus$: element-wise addition or $\otimes$: element-wise multiplication).
The eq.(\ref{eq:att_residual}) can be extended into a more general form -- \textbf{mixture of attentions (MoA)}:
\begin{equation}
\bfs = \underbrace{\bfr \bigcdot \phi(\mathbf{h})}_{\textbf{\small{membership}}}\circledcirc  ~\underbrace{\bA \bigcdot \phi(\bA^\top  \bigcdot \bfx)}_{\textbf{\small{dot-product}}}
\label{eq:att_general}
\end{equation}
 Note eq.(\ref{eq:att_general}) entails the following transformations ($\to$) and captures the representational differences from $\bfr$ to $\bfs$ and from $\bfx$ to $\bfs$:
\begin{align}
\vspace{-3mm}
& \bfx\tikzmark{a} \to \bfr \tikzmark{e} \to \bfp_1 \to \bfs_1 \tikzmark{f} \to \tikzmark{d}\bfs \\
&~~~~~~~~~~~~~\tikzmark{b} \bfp_2 \to  \bfs_2 \tikzmark{c}
  \begin{tikzpicture}[overlay,remember picture]
    \draw[->,black,shorten >=3pt,shorten <=3pt] (a.center) to (b.center);
  \end{tikzpicture}
    \begin{tikzpicture}[overlay,remember picture]
    \draw[->,black] (c) to (d);
  \end{tikzpicture}
\begin{tikzpicture}[overlay,remember picture,out=315,in=225,distance=0.4cm]
    \draw[->,black,shorten >=3pt,shorten <=3pt] (e.center) to (f.center);
  \end{tikzpicture} 
   \vspace{-3mm}
\end{align}
The $\phi(\cdot )$ is either \textsc{softmax}: $p_i=\frac{\exp (z_i)}{\sum_{j}^{k} \exp (z_j)}$ that deterministically computes slice distributions or a Monte-Carlo gradient estimator: \textsc{gumbel-softmax}~\cite{Gumbel:54,jang2016categorical,MaddisonETAL:17}: 
\begin{equation}
p_i = \frac{\exp [(\log(z_i)+\pi_i)/\tau]}{\sum_{j}^{k} \exp [\log(z_j)+\pi_j)/\tau]}
\label{eq:gumbel}
\end{equation}
The $\pi_i$ are i.i.d. samples from the \textsc{Gumbel}$(0, 1)$, that is, $\pi=-\log(-\log(u)), u \sim \textsc{uniform}(0, 1)$. $\tau$ is temperature which controls the concentration of slice distribution, and small $\tau$ leads to more confident prediction over slices. It aims to stochastically compute slice distribution. With Gumbel-softmax, the slice distribution is a \textit{soft} sampling from:
\begin{align}
& \bfp_1 \sim \textsc{Gumbel-Softmax}(\bfh) \\
& \bfp_2 \sim \textsc{Gumbel-Softmax}(\bA^\top  \cdot \bfx)
\end{align}
or a \textit{hard} sampling (but differentiable) from:
\begin{align}
& \bfp_1 \sim \textsc{One-hot}(\arg\max (\bfp_1)) \\
& \bfp_2 \sim \textsc{One-hot}(\arg\max (\bfp_2))
\end{align}
for membership and dot-product attention respectively.

\begin{figure*}[!t]
\centering
\centering
\subfloat{{\includegraphics[width=0.9\textwidth]{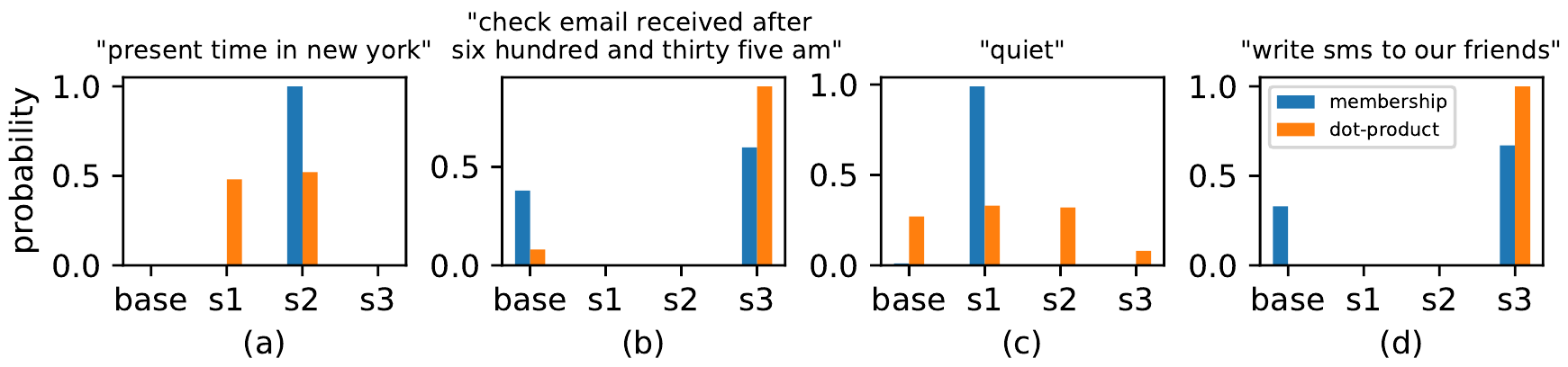}  }}%
\caption{\label{fig:example} Distributions over slices (base, $s1$= Length, $s2$=Time and $s3$=Email) of random test samples that are from membership and dot-product attention mechanisms. In (a)(c) membership attention shows higher confidence while dot-product attention gives higher confidence in (b)(d). Top rows are the utterances from the test set.}
\end{figure*}  
\section{Experiments}
We performed our experiments on a binary classification task with linguistic acceptability and on a multi-class classification task with intent detection.

\begin{table*}[htp]
\centering
\footnotesize
\begin{tabular}{l|ccccc|ccccc}
\hline
 & \multicolumn{3}{c}{F1} & \multicolumn{2}{c}{F1 Lift(\%)} & \multicolumn{3}{|c}{MCC} & \multicolumn{2}{c}{MCC Lift(\%)} \\
Methods & Overall & S1 & S2 & Avg. & Max. & Overall & S1 & S2 & Avg. & Max. \\ \hline 
Baseline & 0.70 & 0.60 & 0.65 & --- & --- & 0.24 & 0.18 & 0.22 & --- & --- \\
SBL & 0.69 & \underline{0.71} & \textbf{0.72} & \textbf{9.0 \%} & \underline{11.0\%} & 0.23 & 0.20 & 0.24 & 2.0\% & 2.0\% \\
SBL-MoA $\oplus$ & 0.70 & \underline{0.71} & 0.68 & 7.0\% & \underline{11.0\%} & 0.25 & 0.24 & 0.20 & 2.0\% & 6.0\% \\
SBL-MoA $\otimes$ & 0.69 & \textbf{0.72} & \underline{0.71} & \textbf{9.0\%} & \textbf{12.0\%} & 0.24 & 0.24 & \underline{0.25} & 4.5\% & 6.0\% \\
SBL-MoA-S $\oplus$ & 0.69 & 0.67 & 0.68 & 5.0\% & 7.0\% & 0.24 & 0.25 & 0.18 & 1.5\% & \underline{7.0\%} \\
SBL-MoA-S $\otimes$ & 0.69 & 0.69 & \underline{0.71} & \underline{7.5\%} & 9.0\% & 0.26 & \textbf{0.28} & 0.22 & 5.0\% & \textbf{10.0\%} \\
SBL-MoA-H $\oplus$ & 0.70 & 0.69 & 0.70 & 7.0\% & 9.0\% & 0.25 & \textbf{0.28} & \underline{0.26} & \textbf{8.0\%} & \textbf{10.0\%} \\
SBL-MoA-H $\otimes$ & 0.69 & 0.69 & 0.65 & 4.5\% & 9.0\% & 0.25 & 0.22 & \textbf{0.32} & \underline{7.0\%} & \textbf{10.0\%} \\ \hline
\end{tabular}
\caption{The results on CoLA test datasets. F1-score and MCC are reported (averaged on 5 random runs for each model). $s1$=Long, and $s2$=Question.The \textit{lift} is the averaged relative improvement across slices over baseline. The largest improvement is in \textbf{bold} and second largest lift number is in \underline{underline} (same to Table \ref{tab:nlu}).}
\label{tab:cola}
\end{table*}

\subsection{Experimental Setup}
\noindent \textbf{Datasets and Metrics}. The CoLA~\cite{warstadt2018neural} dataset has 8551 train and 527 development in domain samples\footnote{\url{https://nyu-mll.github.io/CoLA/}}. We randomly split it into train/val/test with 7200/878/1000 samples. As in \cite{chen2019slice}, we ensure the sample proportion in ground-truth are consistent across splits. We use F1-score and Matthews correlation coefficient (MCC)~\cite{matthews1975comparison} as our metrics. 
The NLU dataset~\cite{liu2019benchmarking} for intent detection contains 25k user utterances across 64 intents. We randomly split it into train/val/test with ratio 0.7:0.1:0.2. We use the accuracy and F1-score as our metrics. 

\begin{table*}[!t]
\centering
\footnotesize
\begin{tabular}{l|cccccc|cccccc} \hline
 & \multicolumn{1}{l}{} & \multicolumn{3}{c} {Acc} & \multicolumn{2}{c|} {Acc Lift(\%)} & \multicolumn{4}{c}{F1} & \multicolumn{2}{c} {F1 Lift(\%)} \\ 
Methods & Overall & S1 & S2 & S3 & Avg. & Max. & Overall & S1 & S2 & S3 & Avg. & Max. \\ \hline
Baseline & 0.7413 & 0.73 & 0.74 & 0.73  & --- &---& 0.7404 & 0.74 & 0.72 & 0.74 & ---  &---\\
SBL & 0.7422 & \underline{0.75} & \underline{0.76} & \underline{0.75} & \underline{2.0\%} & 2.0\% & 0.7418 & 0.74 & 0.72 & \underline{0.75} & 0.3\% & 1.0\%\\
SBL-MoA $\oplus$ & 0.7414 & 0.74 & \textbf{0.77} & 0.74 & 1.7\% & \underline{3.0\%} & 0.7390 & 0.74 & \underline{0.73} & 0.74 & 0.3\% & 1.0\% \\
SBL-MoA $\otimes$ & 0.7440 & \textbf{0.80} & 0.73 & \textbf{0.76} & \textbf{3.0\%} & \textbf{7.0\%}& 0.7411 & \textbf{0.77} & \textbf{0.74} & 0.74 & \underline{1.7\%} & \textbf{3.0\%} \\
SBL-MoA-S $\oplus$ & 0.7403 & 0.74 & 0.75 & 0.73 & 0.7\% & 1.0\% & 0.7403 & 0.73 & \underline{0.73} & \textbf{0.76} & 0.7\% & \underline{2.0\%}\\
SBL-MoA-S $\otimes$ & 0.7424 & \underline{0.75} & 0.72 & 0.75 & 0.7\% & 2.0\% & 0.7421 &0.75 & \textbf{0.74} & 0.74 & 1.0\% & \underline{2.0\%} \\
SBL-MoA-H $\oplus$ & 0.7405 & 0.73 & 0.74 & 0.73 & 0.0\% & 0.0\%& 0.7397 &\underline{0.76} & \textbf{0.74} & \textbf{0.76} & \textbf{2.0\%}  & \underline{2.0\%}\\
SBL-MoA-H $\otimes$ & 0.7418 & 0.74 & 0.73 & 0.75 & 0.7\% & 2.0\% & 0.7401 &0.75 & \textbf{0.74} & 0.74 & 1.0\% & \underline{2.0\%}\\ \hline
\end{tabular}
\caption{The results on intent detection. Accuracy and F1 scores are reported.  $s1$=Length, $s2$=Time, $s3$=Email are the slices that we monitor and aim to improve. The experts' confidence scores are not used as discussed in Sec.\ref{sec:sadrm}.
}
\label{tab:nlu}
\end{table*}

\noindent \textbf{Compared Methods}. We implemented and compared the following methods:
\begin{itemize}
    \item \textbf{Baseline}: A three-layer feed-forward network.
    \item \textbf{SBL}: Slice-based learning~\cite{chen2019slice}.
    \item \textbf{SBL-MoA}: Our approach that extends SBL with a mixture of attentions (MoA).
\end{itemize} 
For SBL-MoA, we developed multiple variants with Gumbel-Softmax. SBL-MoA-S (SBL-MoA-H) are the variant models with soft (hard) sampling from a Gumbel-Softmax distributions. We also tested the way that membership attention and dot-product interact with each other with $\oplus$ (element-wise addition) and $\otimes$ (element-wise multiplication).

\noindent \textbf{Implementation Details}. BERT-base~\cite{devlin2019bert} in sentence-transformer~\cite{thakur-2020-AugSBERT}
is used as the backbone model. We use 128 hidden units for all models, which are implemented with Pytorch~\cite{NEURIPS2019_9015}. A dropout (p=0.5)\footnote{As the data size is relatively small, we use strong dropout regularization to prevent overfitting.} 
is applied after input layer. The models are trained with Adam (0.001)~\cite{kingma2014adam}, with weight decay of 0.01 and 0.001 for the two tasks, respectively. All models are trained with a maximum of 500 epochs with early stopping (patience=50). The best models are selected based on model performance on the validation sets. The temperature $\tau=1.0$ is fixed in all the experiments.

\subsection{Results on Linguistic Acceptability}
\label{sec:linguistic_acceptability}
Table \ref{tab:cola} presents the results on CoLA. First, slice-based models (i.e., SBL, SBL-MoA, and its variants) show that they can maintain (or improve) the original overall performance. Second, we observe that they achieve obvious performance lift on the monitored slices. For instance, SBL achieves an average 9\% F1 score over the baseline. The proposed method (SBL-MoA, $\otimes$) achieves an average of 9\% and maximum 12\% lift.
For MCC, the best performer is SBL-MoA-H, which achieves an average $\geqslant$7\% and maximum 10\% as compared to the baseline. It outperforms SBL by $\geqslant$ 5\%. Also, we notice that using operator $\otimes$ (element-wise multiplication) between the attention mechanisms lead to better performance as compared to $\oplus$.

\subsection{Results on Intent Detection}
\label{sec:intent_classification}

Table \ref{tab:nlu} demonstrates that both SBL and SBL-MoA improve model performance on the monitored slices, with a similar (slightly better) overall performance on the base task\footnote{Note the lift on slice can be negligible to overall due to small size of slice data, e.g., For SBL-MoA $\otimes$, $s1$ with 122 samples, 7.0\% lift only contributes to 122$\times$0.07/5124 $\approx$ 0.0017.}. SBL-MoA variants achieve the best scores and outperform SBL by average 1\% accuracy and 1.7\% F1. 

Figure \ref{fig:example} illustrates the slice distributions given some random samples.
We denote $\bfp_1$ and $\bfp_2$ for membership and dot-product attention respectively. The experiments show that $\bfp_1$ and $\bfp_2$ reach an agreement on predicting the correct slices. 
Interestingly, the sample in (d) --- ``write sms to our friends", in principle, should be sliced as ``base", but both attentions exhibit high confidence to $s3$=``Email". We conjecture the reason is that all utterances are encoded with BERT which captures the similarity between the sample and the utterances in the ``Email" slice.  

\section{Related Work}

SBL~\cite{chen2019slice} is a novel programming model for critical data slices. It is an instance of weakly supervised learning~\cite{zhou2018brief,medlock2007weakly}. The weak supervision data are generated from pre-defined labeling functions~\cite{ratner2016data}. SBL has shown better predictive performance compared to the mixture of experts~\cite{jacobs1991adaptive} and multi-task learning~\cite{caruana1997multitask}, with reduced run-time cost and parameters~\cite{chen2019slice}. The concept of SBL has been recently used in many applications. Penha et al.~\cite{penha2020slice} proposed to adapt SBL to improve ranking performance and capture the failures of the ranker model. Wang et al.~\cite{wang2021handling} recently implemented SBL in a commercial conversational AI system in order to handle the long-tail problem of imbalanced distribution in customer queries and further improved the performance of the conversational skill routing components~\cite{li2021neural,kim2018efficient,kim2018scalable}. 

Our proposed mixture of attention (MoA) is an instance of multi-head attention~\cite{vaswani2017attention} but with different attention types.  MoA can also be extended to include other attention types.
We have shown the effectiveness of this mechanism in determining the slice distributions.

\section{Conclusion}
This paper extends SBL with MoA (SBL-MoA) to improve model performance on particular data slices. We empirically show that SBL-MoA yields better slice level performance lift to baseline and vanilla SBL with two NLU tasks: linguistic acceptability and intent detection. 
\newpage
\clearpage
\bibliography{acl2021}
\bibliographystyle{acl_natbib}

\end{document}